\documentclass[letterpaper, 10 pt, conference]{ieeeconf}  

\IEEEoverridecommandlockouts                              

\usepackage{graphics} 
\usepackage{epsfig} 
\usepackage{amsmath} 
\usepackage{amssymb}  
\usepackage{subfigure}
\usepackage{amsfonts}

\usepackage{amsthm}
\usepackage{gensymb}
\usepackage[utf8]{inputenc}
\usepackage[english]{babel}
\usepackage{color}
\usepackage{mathtools}
\usepackage{algorithm}
\usepackage[noend]{algpseudocode}
\usepackage[font=footnotesize]{caption}
\usepackage{tabularx}
\usepackage{bm}
\usepackage{multirow}
\usepackage{xcolor}
\usepackage{cite}
\usepackage{lipsum}
\usepackage{wasysym}
\usepackage{verbatim}

\include{shortcuts}

\newcommand{\fig}[1]{Fig.~\ref{#1}}

\usepackage[normalem]{ulem}

\title{\LARGE \bf
Towards Transferring Human Preferences from \\Canonical to Actual Assembly Tasks\vspace{-1.0 ex}}

\author{Heramb Nemlekar, Runyu Guan, Guanyang Luo, Satyandra K. Gupta and Stefanos Nikolaidis
\thanks{Heramb Nemlekar, Runyu Guan, Guanyang Luo, Satyandra K. Gupta and Stefanos Nikolaidis are with the University of Southern California, USA.
        {\tt\small \{nemlekar, guanyu, luog, guptask, nikolaid\}@usc.edu}}%
}

\begin{document}

\maketitle
\thispagestyle{empty}
\pagestyle{empty}

\begin{abstract}
        To assist human users according to their individual preference in assembly tasks, robots typically require user demonstrations in the given task. However, providing demonstrations in actual assembly tasks can be tedious and time-consuming. Our thesis is that we can learn the preference of users in actual assembly tasks from their demonstrations in a representative \textit{canonical task}. Inspired by prior work in economy of human movement, we propose to represent user preferences as a linear reward function over abstract task-agnostic features, such as movement and physical and mental effort required by the user. For each user, we learn the weights of the reward function from their demonstrations in a canonical task and use the learned weights to anticipate their actions in the actual assembly task; without any user demonstrations in the actual task. We evaluate our proposed method in a model-airplane assembly study and show that preferences can be effectively transferred from canonical to actual assembly tasks, enabling robots to anticipate user actions.
\end{abstract}


\section{Introduction}\label{sec:intro}

The advent of human-safe robots has enabled deployment of human-robot teams on manual assembly tasks where 
robots can carry out supporting actions, e.g., bringing parts and tools or clearing out the assembly area, while humans focus on the high value actions, e.g., using the tool to assemble the parts. To effectively assist humans, robots need to predict actions that are likely to be performed by humans \cite{hoffman2007effects, lasota2015analyzing, huang2016anticipatory, zanchettin2018prediction}. For example, if a human is expected to perform assembly of a part that requires a screwdriver, the robot can proactively fetch a screwdriver from the tool shelf and deliver it to the human to improve the task efficiency.

However, in many assembly tasks, while a lot of aspects of the task are prespecified and constrained, workers still have their \textit{individualized preferences} on how to execute  a task \cite{nikolaidis2015efficient, munzer2017preference, nemlekar2021two}. For example, one worker may prefer to do all the difficult actions first and easy actions at the end, while another worker may prefer the opposite. Thus, robotic assistants would need to adapt to the individualized preferences of a human operator, e.g., deliver parts in the worker's preferred sequence, to execute the task efficiently and fluently~\cite{grigore2018preference}.

Prior work in learning user preferences for task execution relies on demonstrations (e.g., state-action pairs) of the user in the actual task to learn a policy \cite{argall2009survey, ravichandar2020recent} or an underlying reward function \cite{ziebart2008maximum, palan2019learning, biyik2020learning,puranic2021learning,puranic2021learning2} that captures the user's preferred sequence of actions.
However, providing demonstrations for each assembly task that a given user must perform is especially tedious and time-consuming.

Instead, to reduce the cost of obtaining demonstrations, we posit that we can transfer the user's preference from a representative \textbf{canonical task} (source task) to a new yet related assembly task (target task). We wish our canonical task to be short so that users can easily provide demonstrations, but also expressive enough so that users can demonstrate preferences that enable anticipation of their actions in the actual task.
In this work, we empirically design a canonical task for a given assembly task, and focus on investigating \textit{whether human preferences can be effectively transferred from canonical to actual assembly tasks}. 
Our problem is especially challenging and distinct from prior work in transfer learning \cite{taylor2005behavior,brys2015policy,clavera2017policy}, since we focus on transferring preferences of real users -- as opposed to agent policies -- across tasks.



\begin{figure}[t!]
\centering
\subfigure[Canonical assembly task]{
\includegraphics[width=0.45\linewidth]{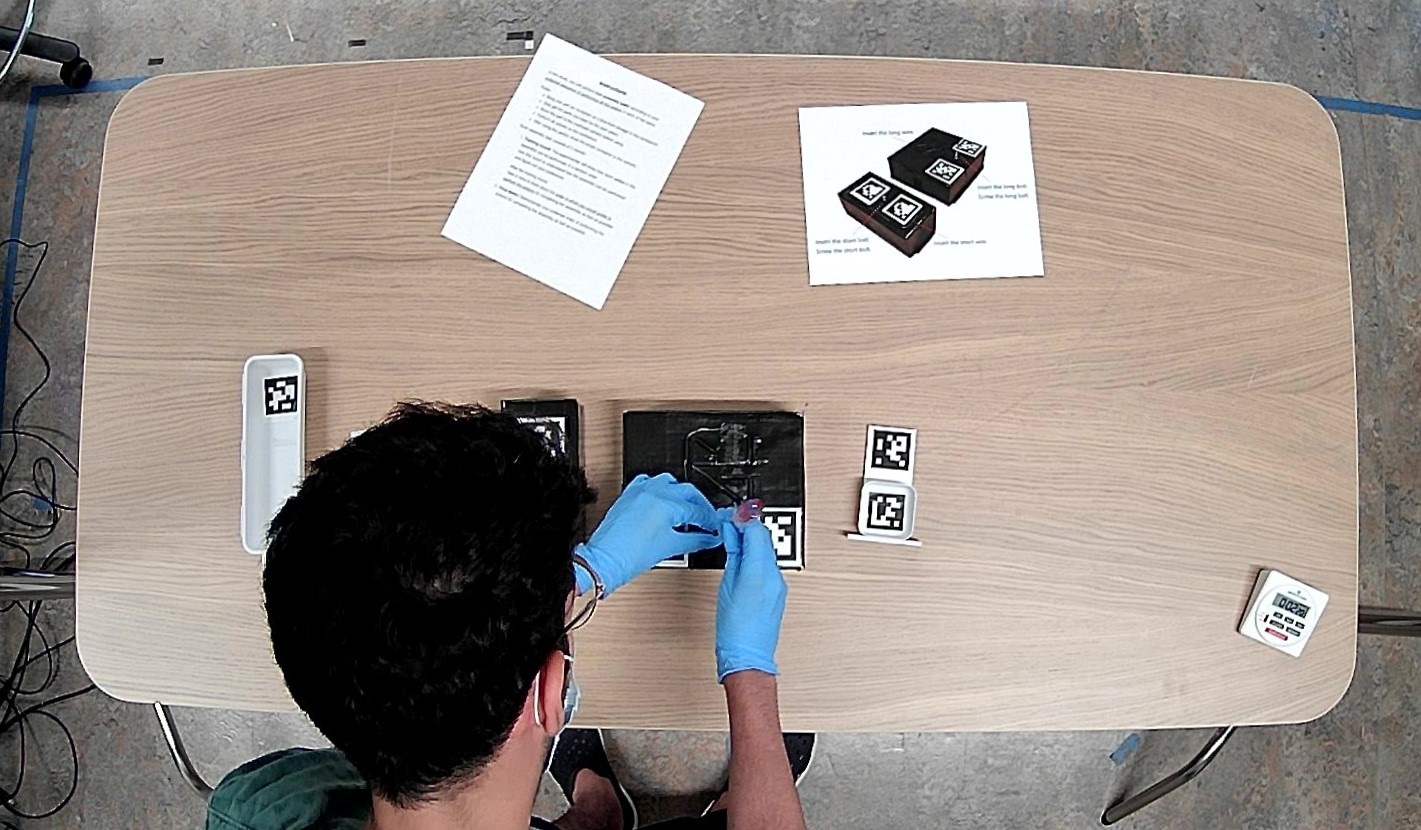}}
\subfigure[Actual assembly task]{
\includegraphics[width=0.45\linewidth]{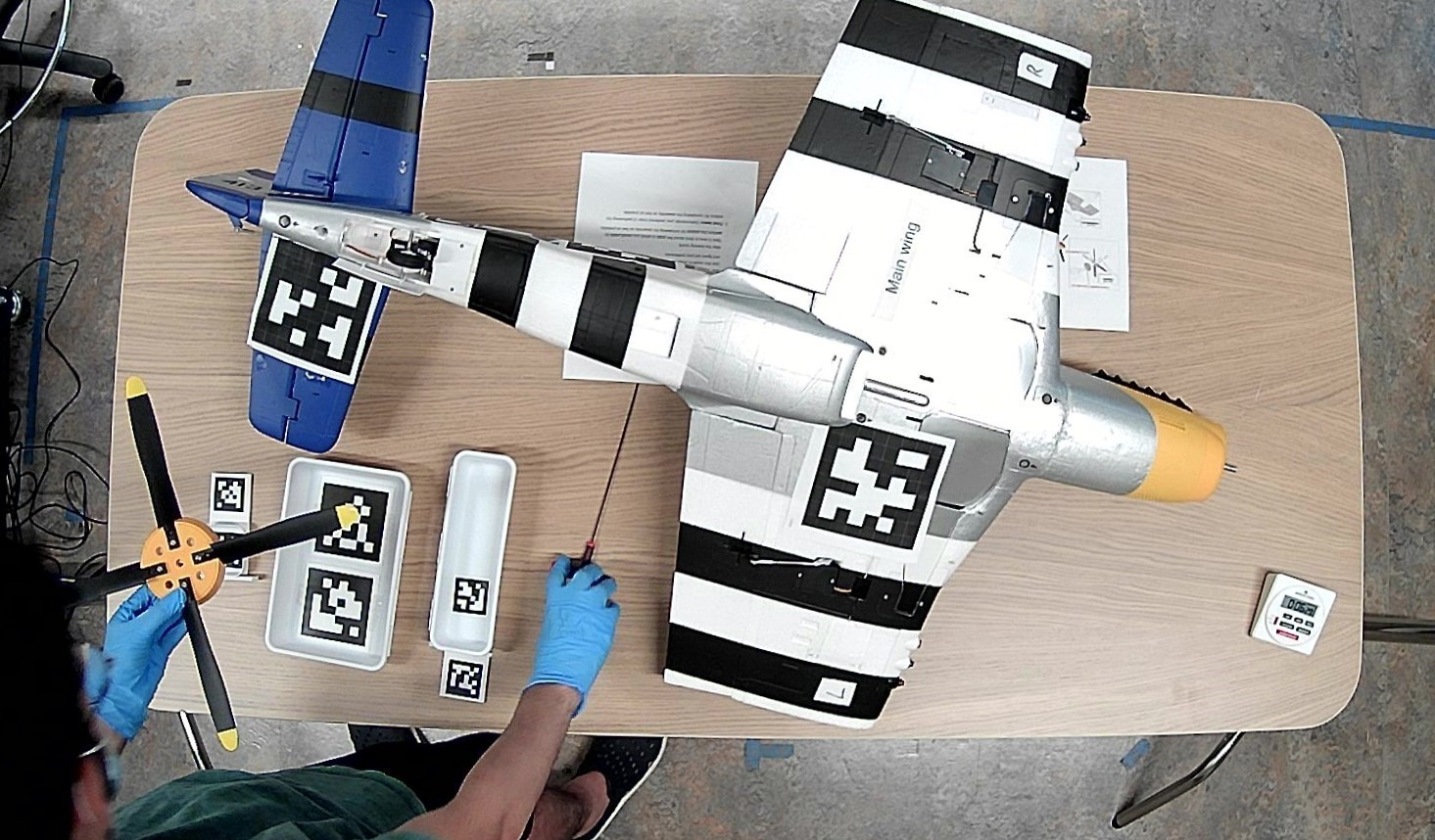}}
\vspace{-1 ex}
\caption{Example of a user that prefers to perform high-effort actions at the end of assembly tasks. (a) Last action of the user in the canonical task is to screw the long bolt which requires the most physical effort. (b) Second last action of user in the actual task is to screw the intricate propeller which also requires the most physical effort (as rated by the user).}
\vspace{-2 ex}
\label{fig:intro}
\end{figure}

Inspired from prior work in economy of human movement \cite{maxwell2001mechanical, zelik2012mechanical, ranganathan2013learning, hesse2020decision} and task ordering \cite{fournier2019task}, 
our key insight is that user preference across different related assembly tasks can be represented with a common set of \textit{abstract,  task-agnostic features}, such as the movement and physical and mental effort required by the user to perform each action in the assembly. Thus, we model the user's internal reward function as linear in the task-agnostic features, where the feature weights represent the user's preference.

For a given user, we hypothesize that their preferences over these features will be similar in both the canonical and actual tasks. For example, if a worker prefers to perform the high-effort actions at the end of the canonical task, they will likely prefer the same in the actual task (see Fig.~\ref{fig:intro}). We use the maximum-entropy inverse reinforcement learning (IRL) \cite{ziebart2008maximum} approach to learn the weights for each user from their demonstration in a canonical task and then use the same weights to model their rewards and compute their policy in the actual task.

\begin{figure*}[t!]
\centering
\includegraphics[width=0.9\textwidth]{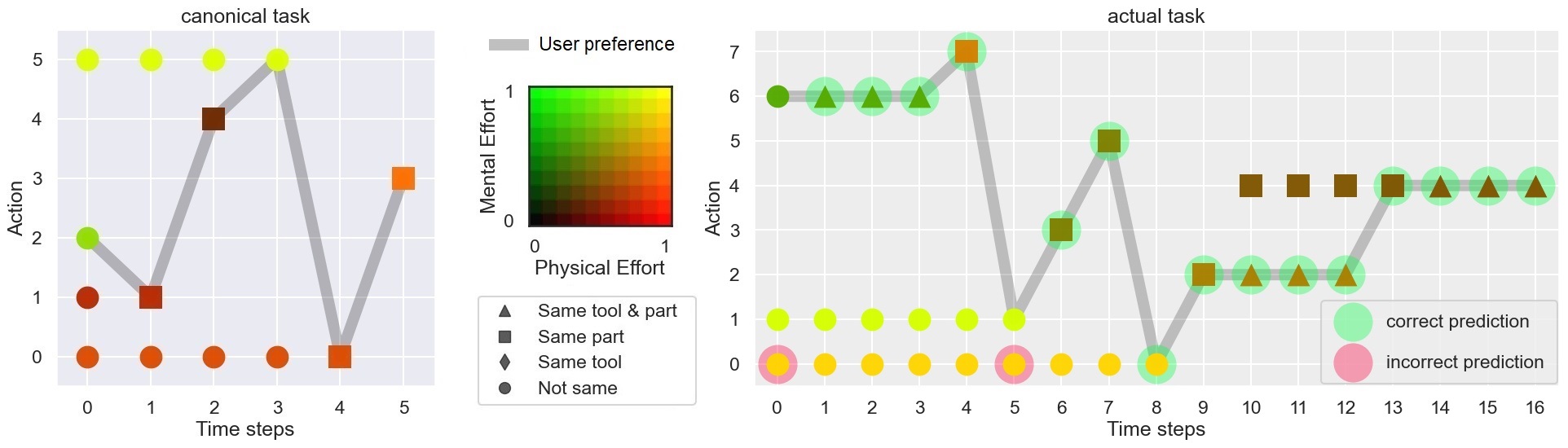}
\vspace{-1 ex}
\caption{Example of a user that prefers to keep working on the same part ($\Square$), and perform actions with high physical effort (red) at the end of the task. In the above plots, the x-axis represents the steps (progress) in the task, and y-axis represents all the different actions in the task. At each time step, the actions that can be executed are marked with a shape that indicates whether that action requires the same tool and part as the previous action (refer to legend). The color of the shape indicates the physical and mental effort required to perform that action (refer to color map). In the canonical task, the user performs action $2$ at the start of the task (time step $0$) because it requires less physical effort (least red) compared to the other choices (actions $0$, $1$, and $5$). At the next time step, the user performs action $1$ because it requires the same part as the previous action $2$. In the remaining steps, the user performs actions on the same part as before, and if no actions use the same part, perform the least physical effort action. Interestingly, the user follows the same preference in the actual task by performing the least physical effort (least red) action $6$ at the start. 
The predictions made by our proposed approach at each time step are shown with a larger green (or red) circle. For further discussion, refer to Section \ref{sec:results}.}
\vspace{-3 ex}
\label{fig:intro_user}
\end{figure*}


Our main contribution is to show how preferences of real users can be transferred from a canonical to an actual assembly task. We validate our proposed method in a user study where we anticipate user actions in a model-airplane assembly task based on their demonstration in a canonical task. Our results show that transferring user preferences from a canonical task can enable accurate anticipation of the user actions in the actual task by modeling user preferences in both the tasks over the same set of task-agnostic features.

\section{Related Work}\label{sec:related}

Similar to prior work \cite{nikolaidis2015efficient, sadigh2017active, palan2019learning}, we consider that the preferences of a user are captured by their internal reward function. Therefore, our goal is to transfer the user's reward function from a canonical task to an actual assembly task.


\subsection{Transferring human preference from source to target task}



The problem of transferring the preferences of real users is distinct from the problem of \textit{transfer learning} \cite{taylor2009transfer, brys2015policy,clavera2017policy} that focuses on using the policies of robotic or simulated agents in a source task as priors to speed-up learning in the target task. Instead, our work focuses on learning the user's preference as a function of abstract features for anticipating their actions in a different assembly task. Moreover, unlike prior work in transfer learning \cite{taylor2005behavior, day2017survey, jing2019task, cao2021transfer}, we attempt to transfer user preferences without access to the user's rewards or demonstrations in the target task.

The prior work most similar to our problem transfers the preference of a simulated human from a (source) block stacking task to a target task with an extra red block~\cite{munzer2017preference} by modeling artificial preferences based on the color of the blocks. However, to our knowledge, no prior work has studied transferring preferences of real users from one assembly task to another.



\subsection{Factors affecting human preferences in physical tasks}

User preferences for sequencing the actions (or sub-tasks) in assembly and manufacturing tasks can be affected by several factors that are task-specific. For example, in a part stacking assembly \cite{wang2018facilitating}, user preferences depended on the size and color of the parts, and were modelled as a linear reward function of the two features. 
While in a Lego model assembly \cite{gombolay2015coordination}, user preferences for task allocation depended on the type of sub-task - fetching or building. 

In order to transfer user preferences from one assembly task to another, we want to model the preferences based on features that are \textit{task-agnostic}. Previous studies \cite{maxwell2001mechanical, ranganathan2013learning, hesse2020decision} have shown that users prefer to minimize movements during task execution. We expect the same for users in an assembly task, i.e., users would prefer to minimize movements required to perform the assembly. 
Similarly, users may also look to minimize their effort spent in the task. For example, in an object pick-up task \cite{fournier2019task} some users preferred to pick up the closest object first because it reduced their cognitive effort, even if it increased their physical effort in the long run. In a study on human jump landing \cite{zelik2012mechanical}, users optimized a combination of active and passive efforts, while also accounting for other factors like safety. 


In this work, we presume that users will prefer to minimize some combination of the movement cost and the physical and mental efforts, with different users having different combinations (i.e. preferences).



\vspace{-0.25 ex}
\section{Methodology}\label{sec:method}

We want to transfer user preferences from a canonical task $C$ to an actual assembly task $X$ for anticipating user actions. We model each task as a Markov Decision Process (MDP) defined by the tuple $(S, A, T, R)$, where $S$ is the set of states in the assembly, $A$ is the set of actions that must be performed to complete the assembly, $T(s_{t+1}|s_{t}, a_{t})$ is the probability of transitioning to state $s_{t+1} \in S$ from state $s_{t} \in S$ by taking action $a_{t} \in A$, and $R(s_{t+1})$ is the reward received by the user in $s_{t+1}$.

We assume that $S_X$, $A_X$ and $T_X$ are known for the actual assembly task $X$. While $T_X$ models the ordering constraints, because each worker can have their own preferred sequence $\xi_{X} = [a_{1}, \ldots, a_{t}, \ldots, a_{N}]$ of performing the actions $a_{t} \in A_X$, $R_X$ will be specific to each worker. Therefore, to anticipate the actions of a user $i$ in the actual assembly task, we must learn their individual reward function $R_{X,i}$.


\textbf{Goal.} We want to learn the user's reward function $R_{X}$ from demonstrations $\xi_C$ provided by the user in a canonical task $C$ (which has its own set of $S_C$, $A_C$ and $T_C$).


\textbf{Intuition.} Our key insight is that preferences of users in actual assembly tasks can be represented with abstract features from a task-agnostic feature space $\Phi$. Given $ \phi: \{S_{C}, S_{X}\} \mapsto \Phi$, we can map any state in the canonical and actual tasks to a $d$-dimensional feature vector in $\Phi$.
   

As the rewards received by a user depend on the state of the task, using $\phi$, we can model the reward function of a given user $i$ as a function of the task-agnostic features.
$$R_{X, i}(s) = f_{X, i}(\phi(s)) \quad \forall \; s \in S_{X}$$

The function $f_{X, i}$ is specific to the user $i$, and captures their individual preference. Our hypothesis is that users will have similar preferences over the abstract features in both the canonical and actual tasks, i.e., $f_{X, i} \simeq f_{C, i}$, given that:
(i) the feature space $\Phi$ fully captures the preferences of all users, and (ii) the canonical task $C$ is expressive enough to capture preferences over a diverse range of feature values.


Knowing $\phi$, we can learn each user's $f_{C, i}$ from their demonstrations $\xi_{C}$ in the canonical task, and use the same function, i.e., $f_{X, i} = f_{C, i}$, to calculate the user's rewards $R_{X, i}$ for states in the actual assembly task.

\begin{figure}[hbt!]
\centering
\vspace{-1 ex}
\includegraphics[width=\linewidth]{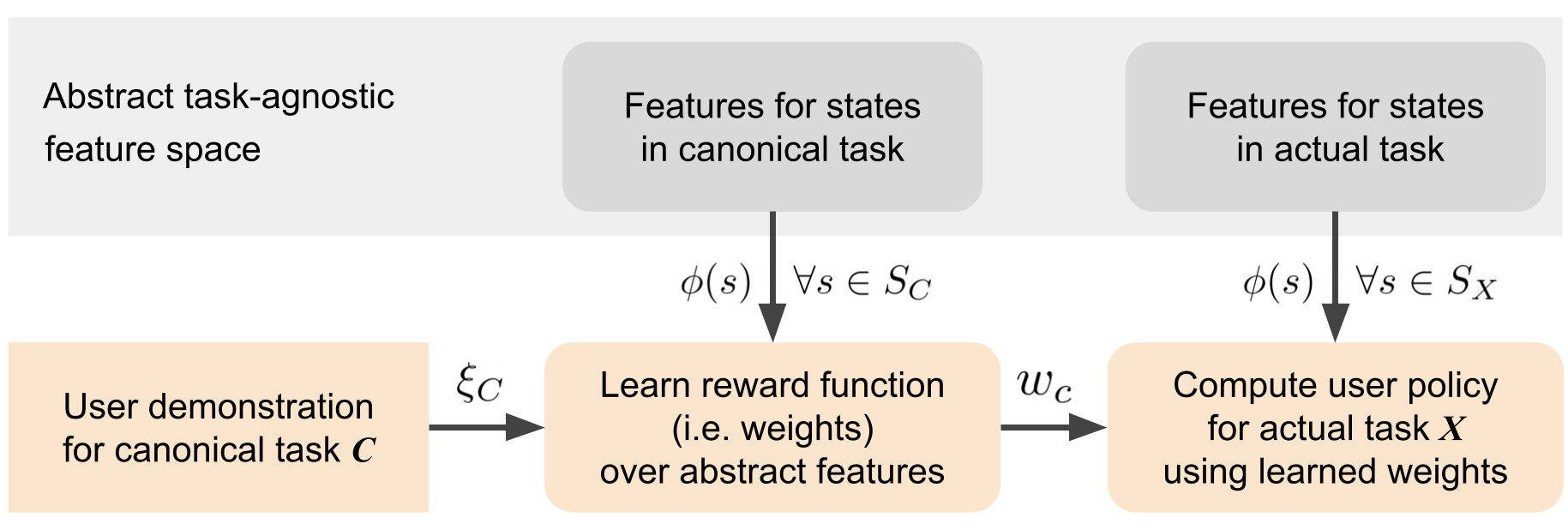}
\caption{Flowchart of our proposed method for transferring preferences}
\vspace{-1 ex}
\label{fig:method}
\end{figure}

\textbf{Approach.} 
Following prior work \cite{ziebart2008maximum}, we model the reward function $R_{X, i}(s) = w_{X, i}^{T} \phi(s)$ as linear in the features $\phi(s)$. Here, $w$ is a $d$-dimensional weight vector where each weight in the vector represents the user's preference for a particular dimension of the feature space.


Given a demonstration sequence $\xi_C = [a_{1}, \ldots, a_{M}]$ of actions $a \in  A_C$, we use maximum-entropy IRL \cite{ziebart2008maximum} to learn the weights $w_C$ for the user. In this approach, we iteratively update the weights to maximize entropy (as there can be multiple solutions) such that our learner visits the features in the canonical task with the same frequency as observed in $\xi_C$. We choose the maximum entropy approach because we wish the learned weights to explain the demonstrations
without adding any additional constraints to the resulting policy. 


Based on our key insight, we assume that user $i$ would have the same weights ($w_{X, i} = w_{C, i}$) for the features $\phi$ in the actual task. Thus, we can calculate the transferred rewards $\tilde{R}_X$ by using the weights $w_C$:  
\begin{equation}
    \tilde{R}_{X, i}(s) = w_{C, i}^{T}\phi(s) \qquad \forall s \in S_{X}
\end{equation}
\fig{fig:method} summarizes our proposed approach for transferring preferences from canonical to actual tasks. We conduct a user study to evaluate whether $\tilde{R}_X$ can be used to effectively anticipate user actions in the actual task.

To anticipate user actions, we assume that the user will try to maximize their long-term reward in the actual assembly task. Thus, we use the learned $\tilde{R}_X$ to perform value iteration \cite{bellman1957markovian} for all states $s \in S_{X}$ in the actual task, and select the action with the highest value as our prediction $\hat{a}_t$ in a given state. In our study, we calculate the value of taking an action in a given state without discounting the future rewards, since users plan for the entire assembly before they demonstrate their preference.\footnote{~For very long assembly tasks users might minimize movement or effort over a shorter horizon, in which case we would adapt the time horizon accordingly.}



\section{Task-Agnostic Feature Space}\label{sec:features}



Inspired from prior work in economy of human movement \cite{maxwell2001mechanical, ranganathan2013learning, hesse2020decision}, we presume that users would try to minimize their movement throughout the task. Because the set of actions is fixed in our assembly tasks, users can minimize their movement when they switch from one action to the next. For example, a worker may prefer to consecutively perform all the actions on one side of the assembly to avoid having to shift sides.
In our user study, movement for switching between actions is performed when the user changes the tool or the part they are working on. Thus, we consider the following features to capture user preferences for minimizing movement:
\begin{table}[h!!!]
\vspace{-1 ex}
\caption{Movement-Based Features}
\vspace{-1 ex}
\label{tab:movement}
\centering
\resizebox{0.75\linewidth}{!}{%
\begin{tabular}{l|l|l|l}
\hline
Feature & Weight & Value & Preference\\ \hline
$\phi_{\mathcal{P}}$ & $w_{\mathcal{P}}$ & $\{0, 1\}$ & Keep same part \\
$\phi_{\mathcal{T}}$ & $w_{\mathcal{T}}$ & $\{0, 1\}$ & Keep same tool  \\ 
\hline
\end{tabular}%
}
\end{table}

Here, $w_{\mathcal{P}}$ and $w_{\mathcal{T}}$ are the weights for the respective movement-based features in the user's reward function. For a state $s_{t+1}$, $\phi_{\mathcal{P}}(s_{t+1}) = 1$ if the latest action $a_{t}$ uses the same part as the previous action $a_{t-1}$. Similarly, $\phi_{\mathcal{T}}(s_{t+1}) = 1$ when $a_{t}$ requires the same tool as $a_{t-1}$. Because we want our features to be state dependent, we augment the current state with the previous two actions: $s_{t+1} \leftarrow [s_{t+1}, a_{t}, a_{t-1}]$. At the start of the task, the previous two actions in the start state are set to $None$.

Previous work~\cite{zelik2012mechanical} also states that users may choose actions trying to minimize the effort they spend in the task.  
For each action, we define the effort $\varepsilon$ as a weighted combination of the nominal physical ($\varepsilon_p$) and mental ($\varepsilon_m$) efforts required to perform that action.
$$\varepsilon(a) = w_{p} \varepsilon_{p}(a) + w_{m} \varepsilon_{m}(a)$$

Here, the weights $w_{p}$ and $w_{m}$ are specific to the user and depend on their individual preference towards minimizing their physical and mental effort.
For example, users may prefer to perform specific actions first to reduce cognitive load, even if it results in requiring more time and physical effort \cite{fournier2019task}. 
Specifically in our pilot studies, we observed that some users preferred performing the high-effort  (mental or physical) actions at the start of the assembly, while others preferred to start with low-effort actions.


If a worker prefers to perform high-effort actions at the end, they must be receiving a higher internal reward for performing the high-effort action at the end instead of at the start. 
We consider this as the \textit{perceived effort} $\varepsilon_{b}$ of an action based on the user's preference to backload the high-effort actions.
To model the time-dependency, we introduce a variable - phase $\psi: s \mapsto [0, 1]$ which represents the percentage of the task that has been completed. We use phase instead of the actual time steps for generality, since the actual task is typically much longer than the canonical task. Using this feature, we model the  perceived effort as a linear 
function of the phase: $\varepsilon_{b}(a, \psi) = \psi \; \varepsilon(a)$.

We can see that at the start, i.e., $\psi	\simeq 0$, the perceived effort will be very small compared to at the end ($\psi	\simeq 1$). Thus to maximize the accumulated reward, users will backload the high-effort actions.
We can also have the opposite scenario, where a workers prefers to perform the high effort actions at the start. In this case, the reward that the user receives for performing the high-effort actions at the start must be higher than at the end. Hence, we model the perceived effort for frontloading high-effort action as: $\varepsilon_{f}(a, \psi) = (1 - \psi) \; \varepsilon(a)$.

As our state contains the information of the latest action, we can model the features for frontloading and backloading actions with high physical effort as:
\begin{equation}
    \phi_{f, p}(s) = \psi_{f}(s)\varepsilon_{p}(s)
\end{equation}
\begin{equation}
    \phi_{b, p}(s) = \psi_{b}(s)\varepsilon_{p}(s)
\end{equation}
Where, $\psi_{f} = 1-\psi$ and $\psi_{b} = \psi$. Similarly, we can calculate the features for sequencing actions based on their mental effort to obtain the following list of features:
\begin{table}[h!!!]
\caption{Effort-Based Features}
\vspace{-1 ex}
\label{tab:effort}
\centering
\resizebox{\linewidth}{!}{%
\begin{tabular}{l|l|l|l}
\hline
Feature & Weight & Equation & Preference\\ \hline
$\phi_{f,p}$ & $w_{f,p}$ & $\psi_{f} \varepsilon_{p}$ &  Frontloading of high $\varepsilon_{p}$ actions \\
$\phi_{f,m}$ & $w_{f,m}$ & $\psi_{f} \varepsilon_{m}$ & Frontloading of high $\varepsilon_{m}$ actions \\
$\phi_{b,p}$ & $w_{b,p}$ & $\psi_{b} \varepsilon_{p}$ & Backloading of high $\varepsilon_{p}$ actions \\
$\phi_{b,m}$ & $w_{b,m}$ & $\psi_{b} \varepsilon_{m}$ & Backloading of high $\varepsilon_{m}$ actions \\
\hline
\end{tabular}%
}
\end{table}

We assume that the effort values for actions in both the canonical and actual assembly tasks can be measured prior to task execution e.g., through user surveys. Therefore, we use the six features from Tables~\ref{tab:movement},~\ref{tab:effort} to create our feature function $\phi(s)$, that maps each state $s$ in the canonical and actual tasks to a $6$-dimensional feature space. 

\section{User Study}\label{sec:study}

We want to show that human preferences learned from demonstrations in a canonical task can be used to anticipate their actions in an actual assembly task. Thus, we conduct a user study where participants demonstrate their preferred sequence of actions in a canonical task and an actual model-airplane assembly task. We use the demonstrations in the actual task as ground truth to measure the accuracy of anticipating actions based on the weights (i.e. preference) learned from demonstrations in the canonical task.

\subsection{Actual assembly task}

\begin{figure}[h!]
\centering
\includegraphics[width=0.53\linewidth]{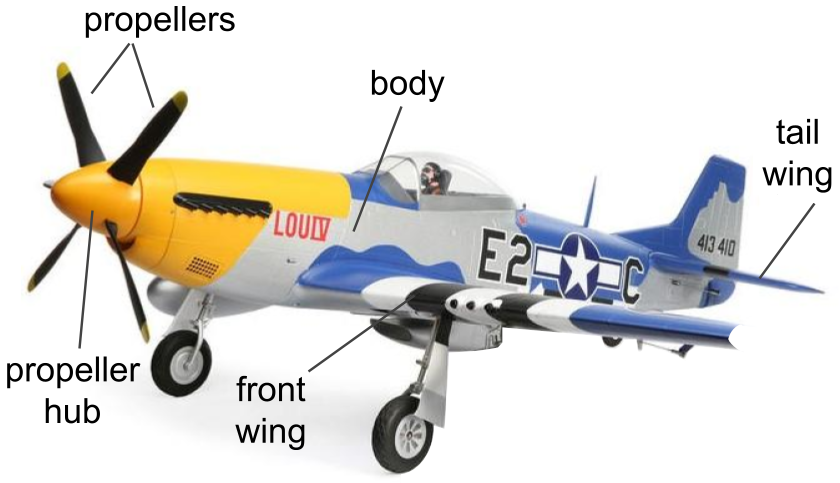}
\quad
\includegraphics[width=0.33\linewidth]{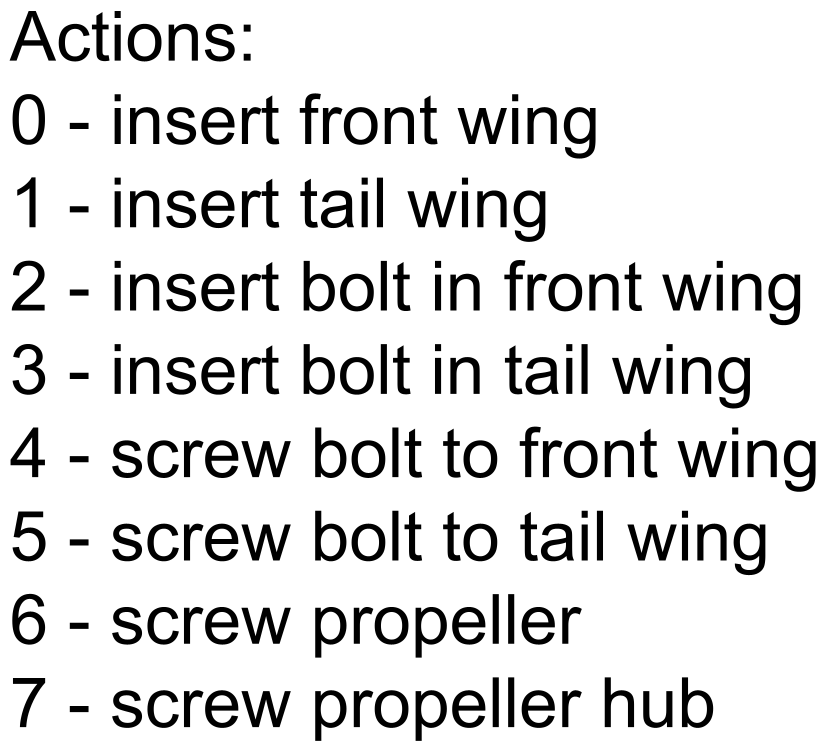}
\caption{Actual assembly task: Airplane model and task actions.}
\vspace{-3 ex}
\label{fig:complex_task}
\end{figure}

We choose an RC model-airplane assembly (see Fig.~\ref{fig:complex_task}) as our actual task. The actions in this assembly task can be sequenced in multiple ways, with few constraints, e.g., action $2$ must precede action $4$ since a bolt must be inserted before it is screwed. The actions can also require different physical and mental efforts. For example, the user shown in Fig. \ref{fig:intro_user} rates the action $0$ of inserting the (large) main wing as having higher physical effort than the action $6$ of screwing a (small) propeller. As some actions need to be repeated, e.g., action $6$ must be performed for each of the 4 propellers, the length of a demonstration in the actual task is $17$ time steps. On average, users required $8.81$ minutes to complete this task.

\subsection{Canonical assembly task}

\begin{figure}[h!]
\centering
\vspace{-2 ex}
\includegraphics[width=0.64\linewidth]{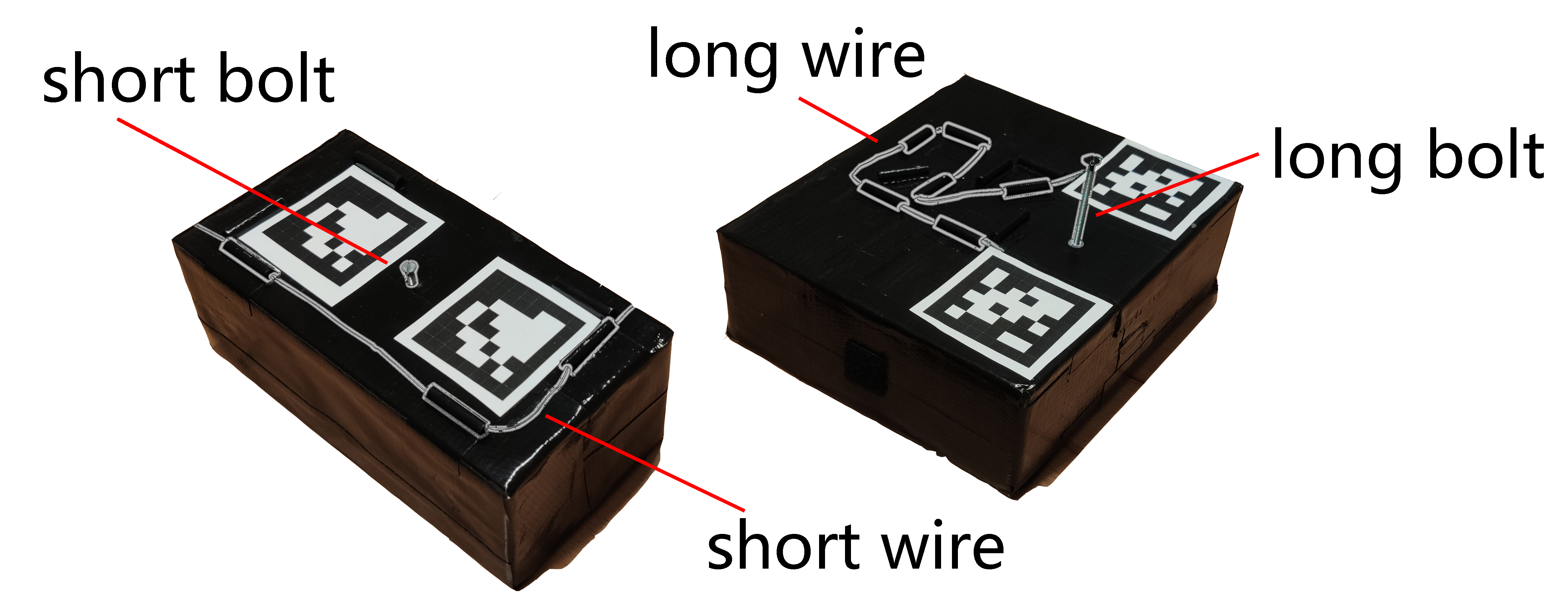}
\includegraphics[width=0.32\linewidth]{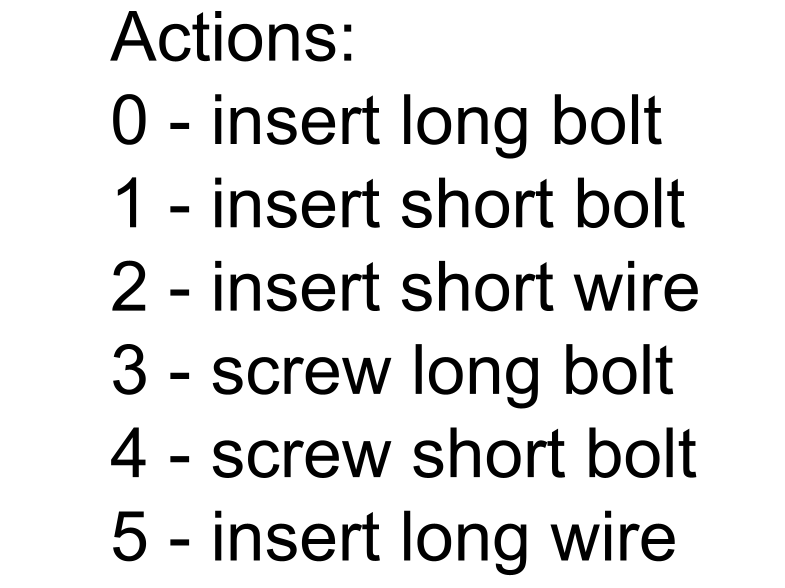}
\caption{Canonical assembly task: (Left) Model, (Right) Task actions.}
\vspace{-1.0 ex}
\label{fig:canonical_task}
\end{figure}

A key challenge in designing the canonical assembly task is to create a subset of diverse states and actions, such that we can capture the user preference over a wide range of feature values, while keeping the task significantly shorter than the actual assembly task.


To capture user preferences for consecutively performing actions related to the same part, we design our canonical assembly with two different parts (see \fig{fig:canonical_task}), each of which is required by at least two different actions.
Next, to capture user preferences for consecutively performing actions that need the same tool, we design a pair of actions that require the same tool, i.e., a screwdriver. Finally, to capture the user preference for sequencing actions based on their physical and mental effort, we design an action for each combination of high and low values for physical and mental effort:
\begin{table}[h!!!]
\centering
\caption{Actions with distinct physical and mental efforts.}
\vspace{-1 ex}
\begin{tabular}{l|l|l}
\hline
 & Low $\varepsilon_p$ & High $\varepsilon_p$ \\ \hline
Low $\varepsilon_m$ & Action 4 & Action 3 \\ \hline
High $\varepsilon_m$ & Action 2 & Action 5 \\
\hline
\end{tabular}
\vspace{-1 ex}
\end{table}

To induce the corresponding physical and mental efforts, we design the actions as follows:
\begin{itemize}
    \item Action $4$, where the user screws a short bolt with a screwdriver. Because the bolt is short, it requires less physical effort to screw it in. Also, since we expect our participants to be familiar with using a screwdriver, we expect this action to require less mental effort.
    
    \item Action $3$, where the user screws a long bolt using a screwdriver. Because the bolt is long, it requires high physical effort to completely screw it in.
    
    \item Action $2$, where the user inserts a wire through three small spacers. Because it requires more focus than the other actions, we expect it to require high mental effort. 
    
    \item Action $5$, where the user inserts a long wire through six spacers. Since there are more spacers, it requires more physical and mental effort to maneuver the long wire.

\end{itemize}

We conducted pilot studies to fine-tune the design of the canonical task and verified that participants perceived the physical and mental efforts of the actions as intended. 
Our final canonical task has $6$ time steps, and on average, users required only $3.83$ minutes to complete the task.
In future, we wish to formalize the process of designing such a canonical task for a given actual assembly task.

\subsection{Study protocol}

We recruited $19$ ($M=12$, $F=7$) participants from the graduate student population at the University of Southern California (USC) and compensated each user with $20$ USD. 

\begin{figure}[hbt!]
\centering
\includegraphics[width=0.42\linewidth]{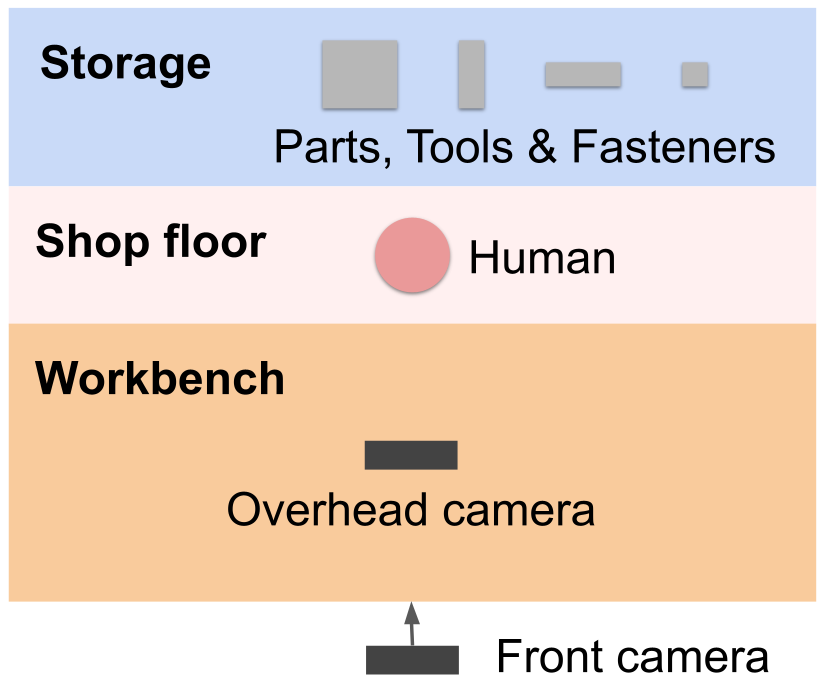}
\;
\includegraphics[width=0.4\linewidth]{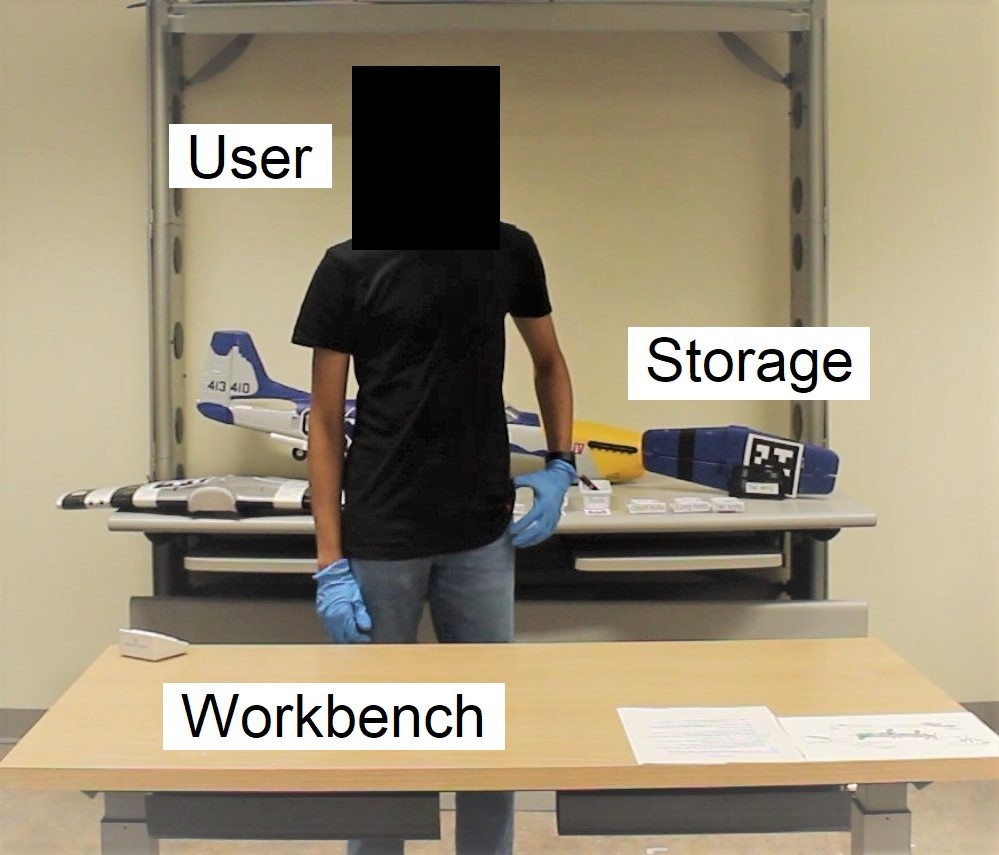}
\caption{(Left) Experimental setup, and (Right) setup at start of actual task}
\vspace{-1.0 ex}
\label{fig:setup}
\end{figure}

\subsubsection{Experimental Setup} 
We divide the space into: (i) a storage area where the parts and tools and fasteners are initially placed, (ii) a workbench upon which the user performs all the actions required to complete the assembly, and (iii) the shop floor where the user can stand and move (see \fig{fig:setup}).
April Tags \cite{olson2011apriltag} are used to track the parts during the assembly with the help of an overhead camera. 

\subsubsection{Study Procedure}
We asked each user to perform both the canonical and the actual assembly task. We counterbalance the order of the tasks to guard against any sequencing effect. For each task, we have a (i) training round - where users learn the assembly task, and an (ii) execution round - where users plan, demonstrate, and explain their preference.

In the training round, we provide each user with a labeled image of the assembled model, as shown in Fig. \ref{fig:complex_task} and \ref{fig:canonical_task}, and we describe all the parts and actions in the assembly. We show how to perform each action in a random order and provide no instruction about the sequence of actions. We also allow users to try each action once for practice, after which they fill in a \textit{post-training questionnaire} to rate the physical and mental effort that they required for performing each action. We obtain the user's values for $\varepsilon_{p}$ and $\varepsilon_{m}$ for each action from these ratings.

In the execution round, we first give users $5$ minutes to plan their preferred sequence of actions for assembling the model as fast as possible. Finally, they demonstrate their preferred sequence and then fill in a \textit{post-execution questionnaire} to report and explain the features that informed their preference in that task. 

\textit{Note:} To avoid influencing the users' preference, we do not inform them that their preference in one task will be used to infer their preference in another and we label the tasks as $A$ and $B$ (not canonical and actual). We also do not tell the users to consider effort or movement while planning their preferred sequence. We simply ask them to demonstrate their preferred sequence of actions in each assembly task. 
\section{Experimental Evaluation}\label{sec:results}


We wish to show that user preferences transferred from the canonical task can be used for action anticipation in the actual assembly task. Our hypothesis is that the accuracy of predicting the users' next action in the actual task based on weights learned in the canonical task would be higher than: (i) randomly picking the next action (\textbf{H1}) and (ii) randomly setting the weights for the features (\textbf{H2}).

We calculate the accuracy of anticipating the users' actions by comparing the action $a_t$ taken by each user in the actual task with the action $\hat{a}_{t}$ predicted by our approach at each time step $t$. The accuracy is $1$ when $a_{t} = \hat{a}_t$, and $0$ otherwise.

\begin{figure}[hbt!]
\centering
\includegraphics[width=0.75\linewidth]{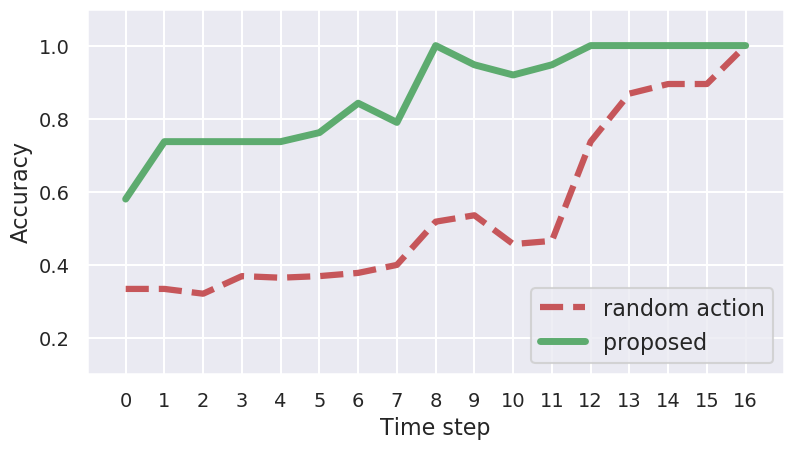}
\vspace{-1 ex}
\caption{Mean accuracy of predicting the user actions at each time step (averaged over all users) in the actual assembly task.}
\vspace{-1 ex}
\label{fig:proposed-vs-random}
\end{figure}

\begin{figure*}[ht!]
\centering
\includegraphics[width=0.9\textwidth]{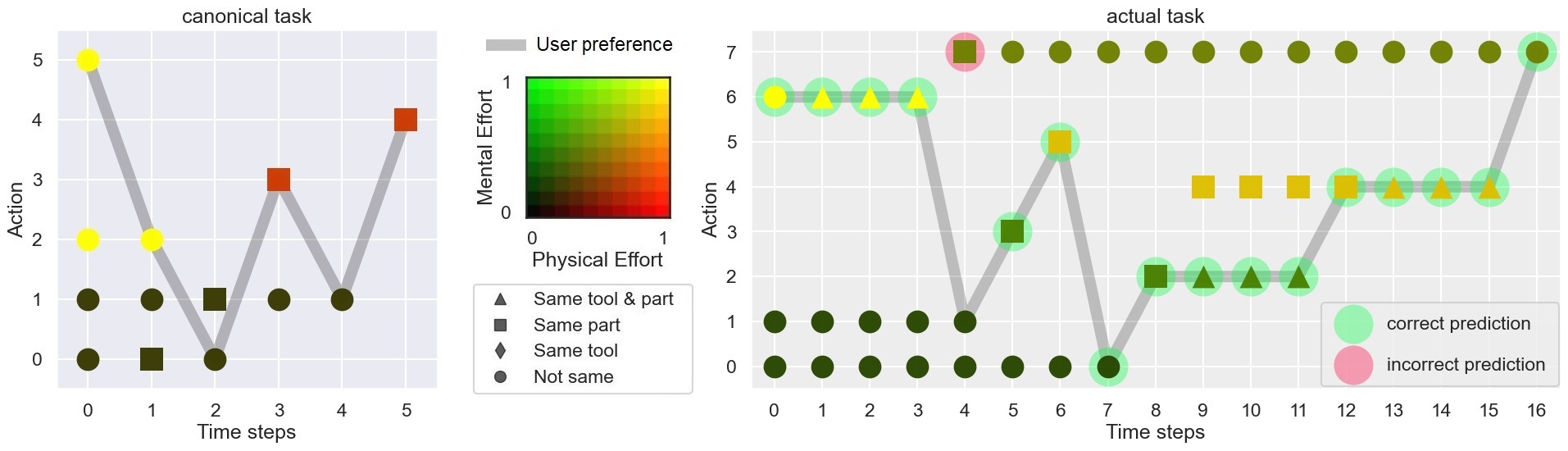}
\vspace{-1 ex}
\caption{Example of a user that prefers to perform actions with the high mental effort at the start of the task. In the canonical task, the user performs actions $5$ and $2$ that have high mental and physical effort (yellow) at the start (time steps 1 and 2). While the user leaves actions $3$ and $4$ that have low mental effort and high physical effort (red) for the end (time steps $3$ and $5$). Thus, from the canonical task demonstration, we learn that the user prefers to frontload actions with high $\varepsilon_{m}$.
In the actual task, the user starts with action $6$ that has the highest mental effort. Based on the canonical task, our proposed method correctly predicts (green circle) that the user will perform action $6$ at time step $0$. Similarly, at time step $5$ the user performs action $3$, which is also what we predict, since it has higher mental effort than actions $0$ and $7$. 
}
\vspace{-3 ex}
\label{fig:good_user}
\end{figure*}

For baseline (i), we randomly select an action from the remaining actions that can be executed at each time step. For each user, we run the baseline for $100$ trials and compute the average accuracy at each time step. We then compare the mean accuracy over all time steps of the proposed method and to randomly selecting actions. A paired t-test showed a statistically significant difference ($t(18)=13.82$, $p< 0.001$) between the mean accuracy for our proposed approach ($M = 0.866$, $SE = 0.032$) and the mean accuracy for random actions ($M = 0.543$, $SE = 0.057$). This supports \textbf{H1}.

In baseline (ii), we calculate $\tilde{R}_{X}$ by uniformly sampling random weights for the actual task features and predict actions based on the computed rewards as in the proposed method. Similar to (i), we run $100$ trials and compute the average accuracy. A paired t-test showed a statistically significant difference ($t(18)=2.93$, $p = 0.008$) between the mean accuracy for our proposed approach and the mean accuracy for random weights ($M = 0.828$, $SE = 0.042$). This supports \textbf{H2}.

Fig.~\ref{fig:proposed-vs-random} shows the mean accuracy of predicting the action at each time step. We can see that the prediction accuracy increases as we reach the final time step, as there are fewer actions to choose at the end. The accuracy for random actions (and random weights) is $0.33$ at the time step $0$ as there are $3$ actions that can be performed at the start. The accuracy for our proposed method is significantly higher at the start as we correctly anticipate the first action for $11$ out of $19$ users based on their transferred weights. 

\section{Discussion}\label{sec:discuss}
We consider two user examples to demonstrate how preferences transfer from our canonical to actual task.

\subsection{Learning user preferences in the canonical task}

Consider the demonstrations shown in \fig{fig:intro_user}, where the user prefers to pick the actions with the same part as the previous action whenever possible (time steps $1$, $2$ and $4$). In the other steps, the user picks the action with the least physical effort (time steps $0$ and $3$). Accordingly, the weights we learn from the canonical task demonstration are higher for the feature of part similarity ($\phi_{\mathcal{P}}$), followed by the feature of backloading high  physical effort actions ($\phi_{b,p}$).
 
On the other hand, consider the user in Fig. \ref{fig:good_user} who prefers to perform actions with high mental effort (actions $5$ and $2$) at the start of the task (time steps $1$ and $2$). Based on their demonstration, we learn a high weight for the feature of frontloading high mental effort actions ($\phi_{f, m}$).


We note that none of the users had the exact same weights $w_C$ even if some users had the same demonstration sequence $\xi_C$, since they gave different ratings for the physical and mental effort of the canonical task actions.

\subsection{Transferring learned preferences to the actual task}

For many users, weights learned for the task-agnostic features in the canonical task enable accurate anticipation of their actions in the actual task.
For the user in Fig. \ref{fig:intro_user}, we learn high weights for minimizing part change ($\phi_{\mathcal{P}}$) and backloading high-effort actions ($\phi_{b, p}$). Using these weights to calculate rewards in the actual task, we can accurately anticipate the actions at $15$ out of the $17$ time steps. For example, we correctly anticipate action $7$ at time step $4$ as it uses the same part as the action $6$ at the previous time step. Similarly, for the user in Fig. \ref{fig:good_user}, we are able to transfer their preference for frontloading actions with high mental effort. For example, at time step $0$, we accurately predict the user's action as $6$, since it has the highest mental effort.

However, in few cases, preferences (weights) over a specific feature did not transfer to the actual task due to human variability. For example, we incorrectly predict at time steps $0$ and $5$ for the user in Fig. \ref{fig:intro_user} since we also learn a high weight for backloading actions with high mental effort ($\phi_{b, m}$) in the canonical task. This is because the last three actions in the user's demonstration have a higher mental effort than the first three actions. Therefore, in the actual task, we predict action $0$ at time step $0$ even if it has high physical effort (low immediate reward), since it would allow the user to perform subsequent low mental effort actions (not shown in the figure) at the start of the task. However, since the user's true preference is to only backload the high physical effort actions, the user performs action $6$ instead. Thus, we see that while the user preference for backloading actions with high physical effort transfers to the actual task, their preference for backloading actions with high mental effort does not.


Finally, we discuss a limitation where the user preference in the actual task was affected by a new feature that wasn't modeled in the canonical task. For example, in Fig. \ref{fig:good_user}, we expect the user to perform action $7$ at time step $4$ based on their preference for frontloading actions with high mental effort. However, we see that the user instead performs action $1$ which has lower mental effort and leaves action $7$ for the end. Based on open-ended responses provided by users at the end of the study, we suspect that the user preferred to leave the action of screwing the propeller hub (action $7$) for the end as they thought that the propellers might break (when they perform the remaining actions) if they were attached at the start. In such cases, we would like the robotic assistant to identify the states where the new feature affects user actions and query the user to actively learn the weights for the new feature. We plan to explore this in future work. 

Overall, we see that preferences can be transferred when the features that determine the user's preferred sequence in the canonical task overlap with the features considered by the user to determine their preference in the actual task.





\section{Conclusion}\label{sec:con}

Our work demonstrates the potential of anticipating user actions in actual assembly tasks based on preferences learned over task-agnostic features in abstract, shorter canonical tasks. Our results show that predictions in the actual assembly task based on transferred preferences are significantly better than the predictions for random actions and random weights. Moreover by learning from user demonstrations in a shorter canonical task we can reduce the time and human effort required to obtain demonstrations in the actual assembly tasks. 

In future, we want to evaluate the benefit of transferring human preferences from canonical to actual assembly tasks by having a robotic assistant perform the anticipated actions.
We would also like to consider longer tasks, where users plan for only a portion of the task and may change their preference over time. While the current setup assumes full observability of the workspace by both the robot and the user, future work will consider sensor placement~\cite{nikolaidis2009optimal} and user viewpoint~\cite{nikolaidis2016based} in robot decision making. Finally, automatically inferring the effort for different actions, instead of using questionnaire responses, is an important area for future work. 

\section*{Acknowledgements}
This work was partially supported by the National Science Foundation NRI (\# 2024936) and the Alpha Foundation (\# AFC820-68). 


\vspace{-1 ex}





\bibliographystyle{IEEEtran}
\bibliography{references}

\end{document}